%% file: main_wacv.tex
\documentclass[10pt,twocolumn,letterpaper]{article}

\usepackage[pagenumbers]{wacv} 

\usepackage{graphicx}
\usepackage{amsmath}
\usepackage{amssymb}
\usepackage{booktabs}

\usepackage[accsupp]{axessibility}  

\usepackage{soul}
\usepackage{transparent}
\usepackage{tabularx}
\usepackage{multirow}
\usepackage{ulem}

\newcolumntype{R}[1]{>{\raggedleft\arraybackslash}p{#1}}
\newcolumntype{L}[1]{>{\raggedright\arraybackslash}p{#1}}

\newcolumntype{Y}{>{\centering\arraybackslash}X}

\usepackage[pagebackref,breaklinks,colorlinks]{hyperref}

\usepackage[capitalize]{cleveref}
\crefname{section}{Sec.}{Secs.}
\Crefname{section}{Section}{Sections}
\Crefname{table}{Table}{Tables}
\crefname{table}{Tab.}{Tabs.}

\renewcommand{\v}[1]{\mathbf{#1}}
\newcommand{\mc}[1]{\multicolumn{1}{c}{#1}}
\newcommand{\formatbest}[1]{\textbf{#1}}

\newcommand{\add}[1]{{#1}}
\newcommand{\remove}[1]{}

\def\wacvPaperID{148} 
\def\confName{WACV}
\def\confYear{2024}

\begin{document}

\title{Recognition of Unseen Bird Species by Learning from Field Guides}

\author{Andr\'es C. Rodr\'iguez$^1$ \footnote{andresro@geod.baug.ethz.ch} 
\and Stefano D'Aronco $^1$
\and Rodrigo Caye Daudt$^1$
\and Jan D. Wegner$^{1,2}$
\and Konrad Schindler$^1$ \\
$^1$ EcoVision Lab - Photogrammetry and Remote Sensing, ETH Zurich, Switzerland
\\
$^2$ Institute for Computational Science, University of Zurich, Switzerland
}
\maketitle

\begin{abstract}
We exploit field guides to learn bird species recognition, in particular zero-shot recognition of unseen species.
Illustrations contained in field guides deliberately focus on discriminative properties of each species, and can serve as side information to transfer knowledge from seen to unseen bird species.
We study two approaches: (1) a contrastive encoding of illustrations, which can be fed into standard zero-shot learning schemes; and (2) a novel method that leverages the fact that illustrations are also images and as such structurally more similar to photographs than other kinds of side information.
Our results show that illustrations from field guides, which are readily available for a wide range of species, are indeed a competitive source of side information for zero-shot learning.
On a subset of the iNaturalist2021 dataset with 749 seen and 739 unseen species, we obtain a classification accuracy of unseen bird species of $12\%$ @top-1and  $38\%$ @top-10,
which shows the potential of field guides for challenging real-world scenarios with many species. \add{Our code is available at \url{https://github.com/ac-rodriguez/zsl_billow}.}
\end{abstract}

\input{1_intro}
\input{2_relatedwork}
\input{3_dataset}

\input{4_methods}
\input{5_experiments}

\input{6_conclusions}

{\small
\bibliographystyle{ieee_fullname}
\bibliography{refs}
}

\clearpage
\input{7_supplementary}

\end{document}

%% file: 1_intro.tex
\section{Introduction}

Fine-grained species recognition is essential for biodiversity monitoring. Identifying the species of observed animals and plants is the basis for several important biodiversity indicators, e.g., the number of different species in an area, the abundance of individual species, and their geographical distribution.
Many species are locally or globally threatened by human activities, making it all the more important to monitor their distributions and support conservation efforts~\cite{ipbes}.

A bottleneck \add{for automatic species recognition in the wild} has long been the collection of enough observations.
\add{There are different modalities for automatic species recognition. Perhaps the two most prominent ones are acoustic recognition from sound recordings and visual recognition form images.
While the focus of this work remains on the latter, acoustic recognition is especially relevant for bird species identification. It is a popular way to do abundance estimation and was explored in early works with Support Vector Machines~\cite{fagerlund2007bird}. Abundance estimation via sound recordings remains an active research area, where new datasets and competitions are being published} \cite{stowell2019automatic,priyadarshani2020wavelet}.
\add{For visual recognition, }in the last years, the cooperation of experts and nature enthusiasts has enabled the emergence of community science projects. Volunteers record and share images
and locations of their observations, which experts can curate and organise to obtain large-scale databases for biodiversity monitoring.
Examples include the iNaturalist~\cite{iNat_web} and eBirds~\cite{sullivan2009ebird} projects. The eBirds platform alone has accumulated \textgreater34 million images for bird species, from $\approx$800'000 contributors.
Those databases make it possible to train automatic species recognition systems, which would be a valuable asset for scalable biodiversity monitoring. 

In principle, automatic species identification can capitalise on the recent advances in computational \add{object recognition. It now achieves human-level performance, and is far more scalable than manual labelling of images; especially in cases where specialized expertise might be needed.}
\footnote{E.g., on ImageNet computers outperform most humans when it comes to recognising different dog breeds, as well as different species of mushrooms.} %

\add{Provided a large volume of labelled training data, one can resort to a supervised learning scheme: A model learns to classify a specific bird species from many images of the bird of interest in many expected natural conditions and backgrounds. This usually means that a large volume of labelled images is needed for training.}
Due to the sheer number of species in most ecosystems, many of which are rare or at least rarely spotted, it can be extremely challenging to gather a sufficient number of training samples for every one of them. For example, the iNaturalist 2021 dataset~\cite{van2021benchmarking} comprises 1'486 bird species, yet the {Birds of the World} collection~\cite{billerman2020birds} reports over 10'000 known bird species.

When data collection is limited, one can resort to machine learning strategies other than supervised learning that may still be able to deliver acceptable recognition results, although these typically do not attain the same performance of a model supervised with enough data.
For instance, one can use 
few-shot learning if only few labelled examples are available for certain classes~\cite{few_shot_survey}. 
In the extreme case, \textbf{Zero-Shot Learning} (ZSL) refers to the scenario where no training samples are available at all for some target classes~\cite{frome2013devise,Lambert2009learning,akata2013label,Yongqin2019theugly}.
This requires class-wise characteristics (side information) rather than labelled data, since labelled examples are not available for training.

\begin{figure}[t]
    \centering
            \includegraphics[width=0.95\linewidth, trim={0 0.5cm 13.7cm 0}, clip]{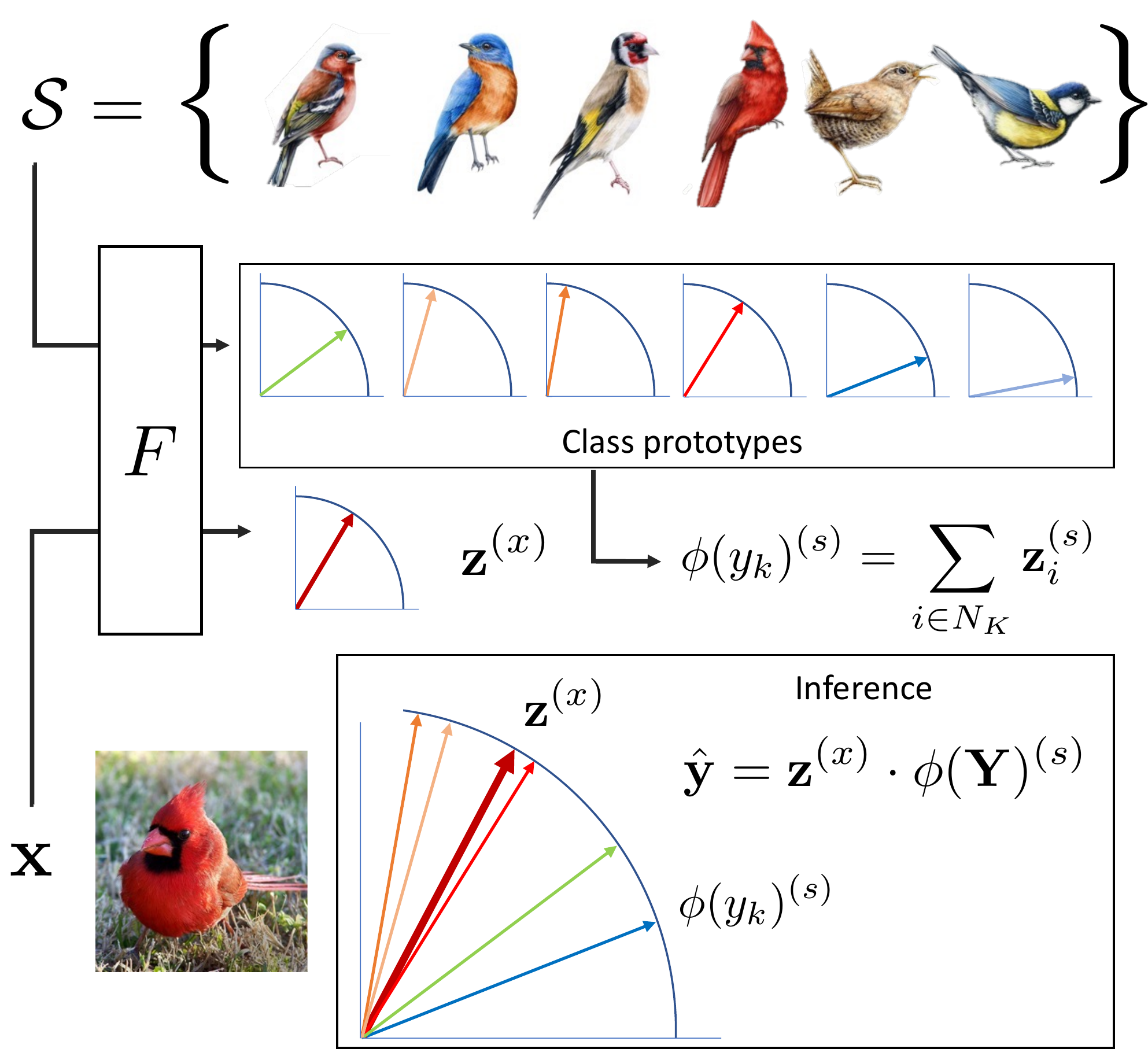}
    \caption{Zero-shot learning with field guides via prototype alignment. Class prototypes (depicted here as different colored vectors) are learned using a shared feature extractor $F$ between photographs and illustrations. At inference time the class with the largest  dot-product to $\v{z}^{(x)}$ is predicted.
}
    \label{fig:teaser}
\end{figure}

Traditionally, professional as well as amateur observers rely on \textbf{field guides} to recognise animal and plant species in nature.
This works remarkably well. Even if new formats of field guides arise, such as interactive maps and mobile apps to aid species recognition~\cite{farnsworth2013next}, the basic principle remains the same: the field guide provides a clear, representative visual example that emphasises the distinctive properties and visual cues needed to identify a species and to discriminate it from similar ones.

The question we explore in this paper is: Can we exploit illustrations from field guides to compensate for the lack of training data for some classes?
Field guides are easily accessible, cover a broad range of species, and although they normally contain only few images of a species -- sometimes only a single illustration -- they allow humans to identify it in most cases.
One can think of a field guide as a collection of manually created, discriminative class prototypes: the artists who create the illustrations are highly specialised professionals, and they make a conscious effort to render each species such that the illustration not only faithfully reproduces attributes like colour and shape, but optimally typifies its peculiarities and makes it distinguishable from other species. Moreover, illustrations are available also for rare, endangered and even for extinct species.

Although naturalistic illustrations resemble photographs in many ways, joint supervised training without additional regularisation leads to biases towards photographic textures, such that the classifier tends to recognise only seen classes.
To tackle this problem, we propose to interpret illustrations as species-specific {attribute} information and leverage them in a zero-shot setting. At this point, a technical difficulty arises: Existing ZSL algorithms ingest attributes in the form of low-dimensional vectors, and we observe that the high dimensionality of illustrations, compared to conventional binary attributes (e.g., belly shape, or eye colour), challenges existing ZSL algorithms. In this work we tackle this problem and demonstrate how illustrations from birding field guides can be exploited for zero-shot learning.

We make the following contributions: (1) We introduce the \textit{Bird Illustrations of the World} (Billow) dataset for fine-grained zero-shot classification of bird species \add{at an unprecedented scale};
(2) we propose a contrastive embedding of the illustrations that enables existing ZSL algorithms to leverage \add{the high-dimensional side information contained in} Billow;
and (3) we propose a novel zero-shot learning scheme better suited for side-information in the form of illustrations. Its fundamental principle is to train a model that can process either illustrations or photographs and in both cases arrives at the same predictions and aligns the class prototypes from the illustrations with the photographs, as depicted in Figure~\ref{fig:teaser}.

\add{We use Billow in conjunction with commonly used datasets with natural images 
commonly used in fine-grained classification.
With the help of those datasets we compare our method to the state-of-the-art in ZSL as well as domain adaptation.}
The experiments show that Billow matches the performance of other, more structured forms of side information, confirming the hypothesis that field guides are a valuable auxiliary source of information for species recognition.
We hope that our work will encourage further research into biodiversity mapping, and may serve as a first step towards unlocking the treasure trove of biological field guides, beyond Billow.

%% file: 2_relatedwork.tex
\section{Related Work}

\textbf{Zero-Shot Learning.}
Early work on ZSL focused on defining class embedding spaces and visual spaces, then measuring some matching metric to predict a class~\cite{frome2013devise,Lambert2009learning,akata2013label}. The embedding space used for matching plays a crucial role~\cite{zhang2017learning,shigeto2015ridge}. Mapping to a space closer to the class embedding can lead to a hubness 
problem, where a classifier is strongly biased to predict only a subset of labels.

Current state-of-the-art methods rely on generative models to map class embeddings into a visual embedding space to avoid such a problem. They use a generator to create synthetic samples that attempt to emulate real samples from unseen classes. These samples are then used to supervise the training of a machine learning algorithm along with the examples from the seen classes.
One of the first studies to use a generative approach in ZSL is \cite{xian2018feature}. They use a Generative Adversarial Network (GAN) to synthesize visual examples of the unseen classes using class descriptions as conditional information. An additional classification loss ensures that the generated features have sufficient discriminative information.
TFVAEGAN~\cite{narayan2020latent}
models the embedding space using a variational formulation. The method also has a feedback network that modulates the latent representations to further improve performance.
Invertible Zero-shot recognition flows~\cite{shen2020invertible} use invertible layers in order to learn a mapping from the class description to the visual features. 
%
Counterfactual ZSL~\cite{yue2021counterfactual} exploits ``sample attributes" from the training classes to create synthetic samples with class attributes from the unseen classes.
%
CE-GZSL~\cite{han2021contrastive} uses a contrastive loss that results not only in class-wise but also instance-wise supervision.
LsrGAN~\cite{vyas2020leveraging} propose a novel semantic regularized loss which promotes visual features that reflect the semantic relationships between seen and unseen classes.
%

\textbf{Side-information in ZSL.} 
ZSL requires a sort of side-information to guide the learning and transfer knowleged from the seen classes to the unseen classes.
The use of illustrations for ZSL is not new. Early work has attempted to use digital characters as side information for character recognition \cite{larochelle2008zero}.
In \cite{antol2014zero}, authors use user generated pose graphics as side information for action recognition in a ZSL setting.
Sketches of objects have been used for image retrieval tasks \cite{sangkloy2016sketchy}.

Generative approaches work very well in cases where the side information has low dimensional 
embedding, and can be used as a deterministic condition by the generator to synthesize samples from unseen classes. 
While there are currently many types of side information used in ZSL, all of them are rather low-dimensional.
%
%
%
%
In \cite{akata2015evaluation}, authors evaluate different supervised and unsupervised embeddings for ZSL. Such types of side information include manually created binary attributes per class~\cite{akata2015label}, 
visual descriptions~\cite{reed2016learning}, automatic embeddings from Wikipedia descriptions~\cite{akata2015evaluation}, and more recently learned embeddings of DNA sequences for fine-grained species classification~\cite{badirli2021fine}.
However, it remains unclear how to use the previously discussed methods if the side information is high-dimensional, as is the case for field guide illustrations, without a low-dimensional embedding step as preprocessing.


\textbf{Domain adaptation.}
Given that illustrations and photograph are similar in nature (as opposed to images and text embeddings or DNA sequences, for instance) we also draw from literature regarding Domain Adaptation (DA). Such studies aim to improve the performance of a model trained on the source domain in which enough data are available for supervision and applied on a target domain in which the available data are not enough for supervision.


The most versatile form of DA is Unsupervised Domain Adaptation (UDA), in which no supervision signal is available for the target domain. Most methods aim to match the distributions of the source and target domains in a latent space, either explicitly or implicitly~\cite{coral,deep_coral,mcc}. \cite{grl} propose using a gradient reversal layer which aims to make the samples from both domains statistically indistinguishable in a representation space. Since, many other methods have been proposed that use adversarial training of a discriminator to enforce alignment of the domains in a latent space~\cite{dann,adda,mdd}.
These methods have been shown to work well in established domain adaptation benchmarks, but use cases for fine-grained classification are somewhat unexplored.

Other studies focus on the case where some supervised samples are available in the target domain. Well established 
methods exist to combine source and target data to improve a ML system's performance on target domain data~\cite{daume2007frustratingly}. \cite{motiian2017CCSA} propose a unified framework for domain adaptation of deep models using a Siamese architecture to align different visual domains.
Notably, \cite{yue2021prototypical,kim2020cross} use memory banks of latent representations of instances from both source and target domains in a few-shot learning setting, which are then used to create class prototypes for each domain.

However, applying these methods directly for ZSL is not straightforward, as seen classes tend to dominate predictions from the target domain if no additional regularization is done. Our novel method aims to close the domain gap from illustrations and photographs in a ZSL setting. 

%% file: 3_dataset.tex
\begin{figure}[t!]
    \centering
\includegraphics[width=0.91\linewidth,trim={0 2.9cm 14.5cm 0},clip]{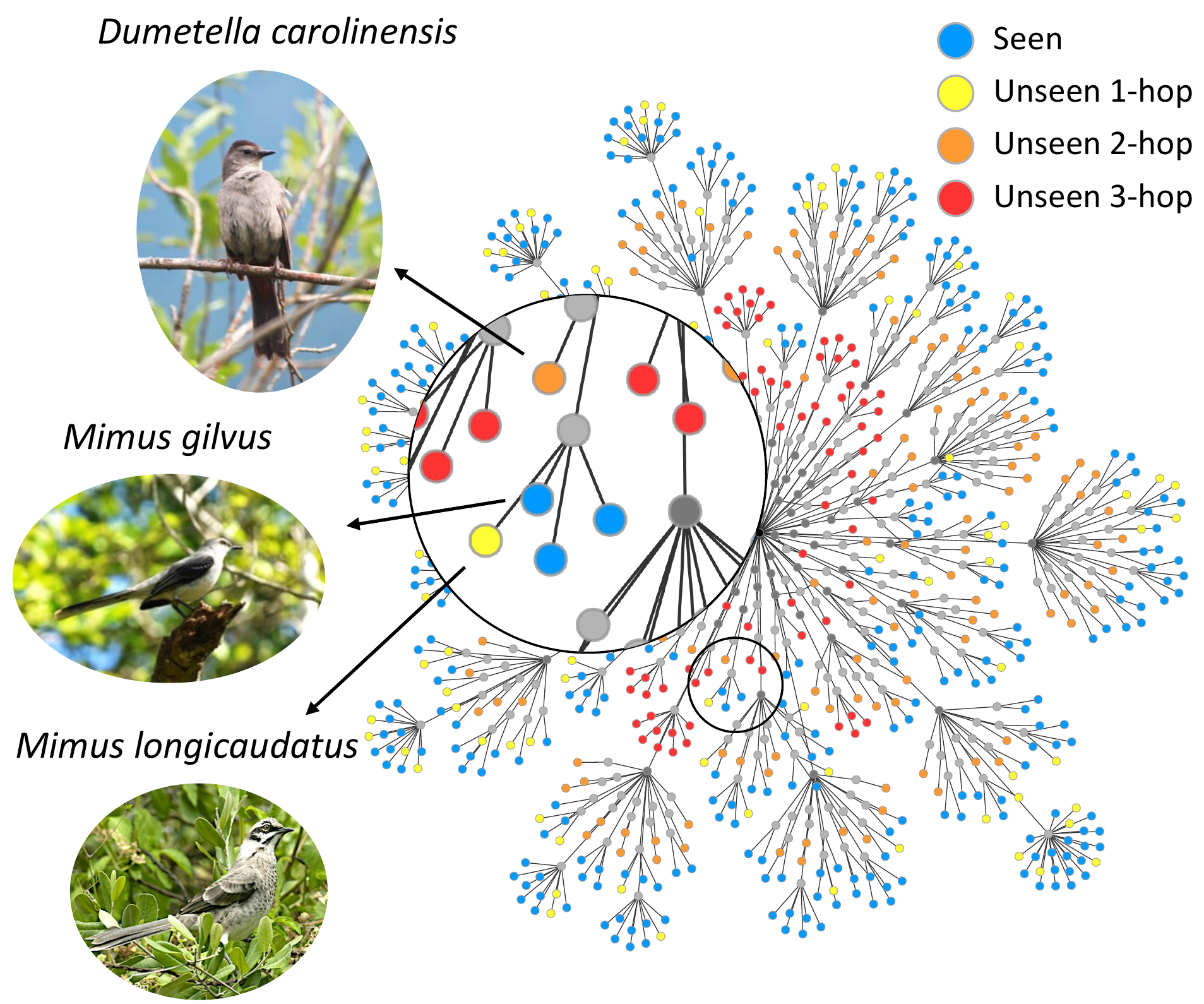}
    \caption{Hierarchical representation of the \emph{Passeriformes} order of the iNat2021 dataset for Seen and $i$-hop unseen classes.}
    \label{fig:iNat2021hops}
\end{figure}

\section{Bird Illustrations of the World Dataset}
\label{sec:dataset}

We introduce the Bird Illustrations of the World (Billow) dataset for Generalized Zero-Shot Learning (GZSL) in fine-grained classification. The dataset consists of illustrations from the {Birds of the World} project~\cite{billerman2020birds} collected and organized by the Cornell Lab of Ornithology.
Billow includes 22'351 illustrations covering 10'631 different species, 2'279 genera, 249 families, and 41 orders.

All illustrations in the dataset share a standardized graphical style: side view in front of white background, in neutral pose. Most species have illustrations for a male and a female specimens, some also include a close-up of the bird's head.

The original artworks may be accessed with a valid subscription to the \textit{Birds of the World} project, and are subject to a licence of use.

For reproducibility and to support further research and comparisons, we describe here how to access all illustrations from Billow. We have used all the images available by November 1st 2021. 
%
%
\add{The encoded dataset, after contrastive embedding with our method, is available along our code on Github. 
To download the raw illustrations we provide a python script. Note though that any further use is subject to the licensing conditions of~\cite{billerman2020birds}.}

\subsection{Illustrations for Zero-Shot Learning}
\label{sec:ilustrations_zsl}
We use Billow with three widely used datasets \add{in Computer Vision}, namely Caltech-UCSD Birds-200-2011 (CUB)~\cite{welinder2010caltech} and the bird subsets of iNaturalist~2017~\cite{van2018inaturalist} and iNaturalist~2021~\cite{van2021benchmarking}.
The list of species included in Billow covers almost all species of the CUB dataset (196 out of 200), and also the overwhelming majority of bird species from iNaturalist 2017 (895 out of 954) and iNaturalist~2021 (1485 out of 1486).
Note that the opposite is not true: even the 1485 bird classes of iNaturalist~2021 are only a small fraction of the 10'631 species present in Billow.
This raises the question of whether we can leverage the rich information contained in the Billow dataset and combine it with a dataset of photographs, to advance the state-of-the-art in fine-grained (bird) species recognition.

For ZSL with CUB, there is a default split into 150 seen and 50 unseen classes~\cite{Yongqin2019theugly}. CUB uses common names, not scientific names. Hence, previous work had to map the common names to scientific ones, e.g., to leverage the hierarchical label structure~\cite{barz2020deep}, or to utilize genetic information~\cite{badirli2021fine}. We have revised and merged these assignments, and only retain mappings for which we found a one-to-one correspondence between the common and scientific name. In Billow we matched 196 out of the 200 CUB classes.

For the iNaturalist datasets, we propose a seen/unseen split. Similar to previous ZSL work that uses ImageNet~\cite{FromeDeViseNIPS2013,norouzi2013zero}, we construct several groups of \textit{unseen} classes, which have different distances to \textit{seen} classes in the label hierarchy. In this way, we can assess the performance of ZSL for unseen classes that are increasingly distant from the seen ones.
%
We first randomly select seen species, and from the remaining species we define the $i$-hop set as the set of all classes whose distance to the nearest seen class in the taxonomic tree is equal to $i$
%
(i.e., they belong to the same super-class at the $i$-th taxonomic level).
For example, the classes in the 2-hop set share the family (2$^\text{nd}$ level) with at least one seen class, but do not share the same genus with any of them.
We consider the species, genus, family and order levels to obtain 0-hop (i.e., seen classes), 1-hop, 2-hop and 3-hop sets. Classes in the 4-hop set do not have members of the same taxonomic group in any level of the seen set. 

The intersection of the \textit{Aves} super-class from iNaturalist 2017 with Billow contains 895 species. These are randomly split into 381 seen and 515 unseen classes. From the unseen ones we construct the 4 different $i$-hop sets for validation.
We repeat the same procedure with iNaturalist 2021: the intersection of its \textit{Birds} super-category with Billow contains 1485 species. These are split into 749 seen and 736 unseen classes. See Fig.~\ref{fig:iNat2021hops} for an illustration of the validation splits, and Tab.~\ref{tab:inat_splits} for the sizes of each split.

\begin{table}[t]
   \resizebox{0.8\textwidth}{!}{\begin{minipage}{\textwidth}
    \include{tables/inat_splits.T}
   \end{minipage}}
    \caption{Zero-shot splits of iNaturalist bird classes. $N$ and $K$ denote the numbers of samples and classes, respectively, in each set}
    \label{tab:inat_splits}
\end{table}

%% file: tables/inat_splits.T.tex
\begin{tabular}{l c *{7}{r}}
    \toprule
\multicolumn{2}{c}{\multirow{2}*{Dataset}} & \multirow{2}*{Train} & \multicolumn{6}{c}{Val} \\\cline{4-9}
        & & & \mc{Seen} & \mc{Unseen} & \mc{1hop} & \mc{2hop} & \mc{3hop} & \mc{4hop} \\
\multicolumn{7}{l}{iNat2017} \\\hline
 &  $N$ &   97,067 &     8,626  & 11,073 &  2,204 &   3,613 &        2,175 &   3,081    \\
 &      $K$ &    381 &  381  & 514 &   87 &       177 &       110 &       140 \\
 \multicolumn{5}{l}{iNat2021} \\\hline
  & $N$ & 211,027 &   7,490 &     7,360 &  1,680 &      2,860 &      1,580 &       1,240 \\
  & $K$ &   749 & 749 & 736 & 168   &  286  &      158 &     124  \\
\bottomrule
\end{tabular}

%% file: 4_methods.tex
\section{Method}
\label{sec:methods}

In ZSL we are given a set of classes $\mathcal{Y}$ made up of two disjoint sets $\mathcal{Y}_{\text{seen}}$ and $\mathcal{Y}_{\text{unseen}}$.
Side information $\v{s}$ is available for every class in $\mathcal{Y}$. In most cases a single instance of $\v{s}$ is available for each class, although this is not a requirement.
We can think of $\v{s}_y$ a (possibly incomplete) description of the class $y \in \mathcal{Y}$, e.g., a text, a list of semantic attributes, or a gene sequence.
A training set of pairs $(\v{x}, y)$ is available exclusively for the seen classes. In computer vision, $\v{x}$ typically refers to photographic images.
The goal of ZSL is to use that information to build a classifier $F(\v{x})$, which can recognize samples from \textit{unseen} classes $\mathcal{Y}_{\text{unseen}}$. Similarly, Generalized Zero-Shot Learning (GZSL) methods aim for good classification performance on test samples from both the seen \textit{and} unseen classes $\mathcal{Y} =  \mathcal{Y}_{\text{seen}} \cup \mathcal{Y}_{\text{unseen}}$.

The side information $\v{s}$ provides cues about similarities (common features) between classes and serves as a bridge that enables the recognition of instances of unseen classes.
Commonly the side information comes in the form of low-dimensional vectors, like for instance binary presence/absence flags for a number of attributes. On the contrary, our illustrations are images of a specific style / domain.

In order to utilise these illustration for ZSL, we explore two different strategies. We start with a \add{two-stage strategy}, where we first learn a \add{ \textit{Contrastive Encoding} of the illustrations}, such that the resulting codes can be fed into existing ZSL methods \add{at a second stage}.
We then go on to develop a more advanced method, \add{named \textit{Prototype Alignment}}, where a single \add{end-to-end} network is trained to map both illustrations and photographs to similar latent representations, in order to better leverage their similar structure. 

\subsection{Contrastive Encoding of Billow}
\label{sec:vetor_encoding}

\add{
Standard ZSL methods in literature assume the existence of class descriptions $\phi(y)$ in the form of low-dimensional vectors derived from some sort of side information that should help recognise seen and unseen classes. We must therefore compute $\phi(y)$ as a first step to use those methods.} 

Let $\mathcal{D}_{\text{side}}$ be the set of pairs $(\v{s}, y)$ where $\v{s} \in \mathcal{S}$ is an illustration associated with class $y \in \mathcal{Y}$. To turn illustrations into low-dimensional vectors, we use an encoding network $E$ that produces an embedding $\v{z} = E(\v{s})$. These embeddings should preserve discriminative class information, so we simply add a classification head $\hat{\v{y}} = C(\v{z})$ 
and optimize $E$ and $C$ with a cross-entropy loss $L_{\text{cls}}(\hat{\v{y}}_{i}, \v{y}_i)$. Here, $\v{y}$ represents the one-hot encoding of $y$.

However, the ability to discriminate classes is not enough. The embeddings should also live in a metric space where pairwise differences between them are meaningful, so as to handle the zero-shot setting.
%
One way to achieve this is to model the overall distribution of illustrations in an embedding space. For instance, one can employ a Variational Auto-encoder (VAE), that assumes the embedding $\mathbf{z}$ to model a prior distribution in the latent space from which it is possible to draw samples and decode them to the original input space $\mathcal{X}$ \cite{Higgins2017betaVAELB,burgess2018understanding}. 
This approach, however, risks reducing the representation power of $\v{z}$ if it is too strongly regularized by the prior distribution 
(see Supplementary).

An alternative is to use a contrastive loss that promotes an embedding space with a uniform distribution over
the unit-sphere
\cite{Prannay2020supervcontr}.
We apply a projection and normalization head $\Tilde{\mathbf{z}} = h(\mathbf{z})$ 
to the embeddings before computing the contrastive loss. Following \cite{Prannay2020supervcontr}, our contrastive loss function is
\begin{equation}\small
     L_{\text{cont}} (\Tilde{\v{z}}_{i}) = - \dfrac{1}{||P(i)||_1} \sum_{p \in P(i)} \log \frac{\exp\left(\frac{1}{\tau}\Tilde{\v{z}}_{i}\Tilde{\v{z}}_{p}\right)}{\sum_{j  \in B} \exp\left(\frac{1}{\tau}\Tilde{\v{z}}_{i}\Tilde{\v{z}}_{j}\right)},
     \label{eq:contrastive_loss_encoded}
\end{equation}
where $P(i)$ 
is a set of samples in the training batch $B$ that have the same class label as $\v{x}_{i}$, and $\tau \in \mathbb{R}^+$ is a tunable temperature parameter.
Finally we train $F$, $C$ and $h$ with both classification and contrastive losses:
$L = L_{\text{cls}} + L_{\text{cont}}$. 
%
%
As final representation of class $y$, we compute
\begin{equation}\small
    \phi (y) = \eta \left(\sum_{\v{s} \in \mathcal{S}_{y}} E(\v{s})\right) ,
\label{eq:descriptor_naive_method}
\end{equation}
where $\mathcal{S}_{y}$ is the set of all illustrations available for class $y$, and $\eta(\v{z}) = {\v{z}}/{ ||\v{z}||}$ denotes $L^2$-normalization. The embeddings $\phi(y)$ derived from illustrations can then be used as class descriptors in different existing ZSL methods.

\subsection{Prototype Alignment}

In contrast to other types of side information for ZSL, illustrations also belong to the visual domain. We leverage this property and propose Prototype Alignment (PA) for ZSL with visual side information\add{, which allows us to bypass the encoding step required by all previous ZSL-methods.}
Inspired by~\cite{yue2021prototypical}, 
we explore a view of the problem through the lens of few-shot \textit{domain adaptation}: The source domain are illustrations, the target domain are natural, photographic images.

Let $\v{s}$ and $\v{x}$ be samples from the source domain $\mathcal{S}$ and the target domain $\mathcal{X}$, respectively. We have access to samples from all classes $\mathcal{Y}$ in the source domain, but only to samples of the seen classes $\mathcal{Y}_{\text{seen}}$ in the target domain. Furthermore, we also do not have unlabelled samples of unseen classes in the target domain.

We train a feature extractor network $F$ that takes input samples from either domain and outputs a latent representation $\v{z}$. The last operation in $F$ is an $L^2$-normalization layer $\eta(\cdot)$, as also used in Eq.~\eqref{eq:descriptor_naive_method}.
%
During training, we keep a memory bank in each domain, with a prototype $\v{z}$ of each class. For the illustrations in the source domain, that representation can be interpreted as the class embedding $\phi(y_k)^{(s)}$ that is used for ZSL. %
Note that, in contrast to previous approaches~\cite{kim2020cross,yue2021prototypical}, we do not keep an instance-wise memory bank, which would lead to intractable memory demands for larger datasets.

For the sake of simplicity, we omit the domain indicator from this point on where possible.
In every iteration, we update the memory bank in each domain with the latent representation of the new samples, with momentum $m$:
\begin{align}
\phi(y_k) & \gets  \eta\left((1-m) {\v{z}_{k}} + m \phi(y_k)\right).
\label{eq:momentum_centroid}
\end{align}
%
%
To promote compact and discriminative class representations, we apply a contrastive in-domain loss similar to Eq.~\ref{eq:contrastive_loss_encoded}, via a projection head $h$:
\begin{equation}\small
  L_{c} \big(\v{z}_{i}, \phi(y_i)\big) = -\log\frac{\exp\left(\frac{1}{\tau}h(\v{z}_{i}) h(\phi(y_i))\right)}{\sum_{k \in C} \exp\left(\frac{1}{\tau}h(\v{z}_{i})h(\phi(y_{k}))\right)}.
\end{equation}
%
%
In contrast to \cite{yue2021prototypical} we refrain from applying a cross-domain contrastive loss to close the domain gap. Instead, we sidestep the gap by directly using the class prototypes from \textit{both} domains for classification, so as to force the network $F$ to produce class-discriminative features.
%
To obtain class logits, we compute the dot-product between an image embedding $\v{z}$ and the embeddings $\phi(\mathcal{Y})$ 
of the classes from both domains, $\hat{\v{y}}^{(s)} =  \v{z} \cdot \phi(\v{Y})^{(s)} $ and $\hat{\v{y}}^{(x)} = \v{z} \cdot \phi(\v{Y}_{\text{seen}})^{(x)} $.
%
These serve as input to a cross-entropy loss $L_{\text{cls}}$
for supervision: 
\begin{equation}\small
L_{\text{cls}}\left(\hat{\v{y}}^{(s)},\hat{\v{y}}^{(x)}, \v{y}\right) = L_{\text{cls}}\left(\hat{\v{y}}^{(s)}, \v{y}\right) + L_{\text{cls}}\left(\hat{\v{y}}^{(x)}, \v{y}\right).
\label{eq:ce_da_method}
\end{equation}
Eq.~\ref{eq:ce_da_method} 
encourages sample representations that are discriminative w.r.t.\ prototypes from the \textit{other} domain, which in turn aligns the two domains.
Note also that the second term in Eq~\ref{eq:ce_da_method} is only computed for seen classes, as it depends on $\phi(\v{Y}_{\text{seen}})^{(x)}$.
%
%
The complete loss function is $L =L^{(s)} + L^{(x)}$, such that
\begin{equation}\small
     L^{(d)}\!=\!\!\!\sum_{i \in B^{(d)}}\!\!\Big(\!\lambda_{c}^{(d)} L_c(\v{z}_{i}, \phi(y_i))
     \!+\!\lambda_{\text{cls}}^{(d)} L_{\text{cls}} (\hat{\v{y}}_i^{(s)},\hat{\v{y}}_i^{(x)}, \v{y}_i)\!\Big),
\end{equation}
where $B^{(d)}$ denotes indices of the samples from domain $d \in \{\mathcal{S}, \mathcal{X}\}$ in the mini-batch. Hyperparameters $\lambda_\text{c},\lambda_{\text{cls}}$ are used to balance the different losses. At test time, we can simply use the logits $\hat{\v{y}} = F(\v{x}) \cdot \phi(\v{Y})^{(s)}$ for classification.

%% file: 5_experiments.tex
\begin{table*}[ht]
    \begin{subtable}[t]{\linewidth}
    \caption{Seen (S), unseen (U) and harmonic mean (H) top-$k$ accuracy. Average of 5 runs $\pm$ standard deviation.}
       \centering
      \resizebox{0.86\linewidth}{!}{\begin{minipage}{\linewidth}
        \include{tables/inat_suh_topk}   \end{minipage}}
        \label{tab:inat_suh_topk}
    \end{subtable}
    \begin{subtable}[t]{\linewidth}
    \caption{Unseen $n$-hop validation sets top-$k$ accuracy. Average of 5 runs. }
      \centering
   \resizebox{0.86\linewidth}{!}{\begin{minipage}{\linewidth} \centering
        \include{tables/inat_nhops_topk}   
        \end{minipage}}
        \label{tab:inat_nhops_topk}
    \end{subtable}
    \caption{GZSL on iNaturalist Datasets with Billow. \underline{CE:} Contrastive Encoding of illustrations and TFVAEGAN. \underline{PA:}  Prototype Alignment. Best method is marked in {bold}.
    }
    \label{tab:inat_results}
\end{table*}

\section{Experiments}

\textbf{Experimental Setup.} 
All of our experiments are developed using PyTorch~\cite{paszke2017automatic} and trained with Nvidia GTX 1080 GPUs.
For our contrastive encoding of illustrations we use a ResNet-18~\cite{He_2016_CVPR} pretrained on ImageNet to create the embeddings $\phi(y)$ from the illustrations. As is commonly done in ZSL literature, features from a pretrained ResNet-101 backbone without fine-tuning were used to obtain a 2048-dimensional vector representation of each image.

The PA experiments used a {ResNet-101} pretrained on ImageNet data and used the Adam optimizer~\cite{kingma2014adam} with a base learning rate of $10^{-4}$, and the convolutional layers' learning rate scaled down by $0.1$. {All experiments on the iNaturalist datasets ran for 40.000 iterations and experiments on CUB for 200 epochs}. We set $\tau = 0.1$ in all our experiments. 
\add{We retrained all baselines using their respective original implementations.}

Following the convention in GZSL literature, we evaluate the performance of each algorithm using held out sets of samples of the seen classes (S) and unseen classes (U) separately. The harmonic mean of these two numbers (H) is also reported.
%
We will make our all our code available for reproducibility.

\subsection{Zero-Shot Recognition, iNaturalist 2017 and 2021}
\label{sec:zsl_inat}

\add{We introduce the first results with ZSL leveraging the illustrations from Billow and the images from iNaturalist datasets.} We report experiments using CE with TFVAEGAN~\cite{narayan2020latent} in a two-stage approach, and experiments using Billow illustrations directly with PA. 
{
On all iNaturalist datasets we observed an improved performance of PA over the CE. This was consistent on all three datasets evaluated on all top-$k$ metrics.
With PA we observed a harmonic mean H@top-5 of 35.1\% and 35.6\% for iNat2021 and its iNat2021mini, respectively (see Tab.~\ref{tab:inat_suh_topk}). For CE we observed a decreased performance with the larger training dataset for iNat2021 (\mbox{H@top-5} 19.1\% and 24.6\%).
%
{These results indicate that further regularization may be needed for large datasets.}
}


Table~\ref{tab:inat_nhops_topk} shows that the hierarchical distance to the nearest seen classes correlates strongly with performance on the unseen datasets.
\add{Similar as previously observed, CE had a decreased performance with respect to PA. This was consistent over all $i$-hop sets. We also evaluated performance at different hierarchy levels and found a similar behaviour. See the Appendix for details and further analysis.}
\remove{We observe a constant reduction on performance for classes with a larger taxonomic separation to seen classes.}
%
This is aligned with what has been observed in ImageNet for ZSL~\cite{FromeDeViseNIPS2013,norouzi2013zero,kampffmeyer2019rethinking}. However, it seems that ZSL on ImageNet is more challenging than for iNaturalist, perhaps because the label distances between ImageNet classes are not as meaningful as taxonomic distances between species.

\textbf{Analysis of the number of synthetic samples.}
Generative approaches in GZSL generate synthetic examples for unseen classes, which is usually set to $N_{\text{syn}} = 100$ in works CUB~\cite{narayan2020latent,han2021contrastive,vyas2020leveraging}. The samples are added to the training set to supervise the training of a classifier. We investigate the effect that bigger values of $N_{\text{syn}}$ could have on larger datasets with higher numbers of unseen classes. We kept all other hyperparameters constant. 
Results can be found in Table~\ref{tab:inat_synnum_topk}. 
We observe that with TFVAEGAN increasing $N_{\text{syn}}$ results in better performance for unseen classes in all the datasets, but at the cost of lower performance for seen classes.

\begin{table}[ht]
     \resizebox{0.75\linewidth}{!}{\begin{minipage}{\linewidth}
       \centering
        \include{tables/inat_synnum_topk}  
       \end{minipage}}
 \caption{GZSL on iNaturalist Datasets with Billow. Results with Contrastive Encoding and TFVAEGAN  with different number of synthetic samples. Average of 5 runs, best method is {bold}. Seen, unseen and harmonic mean (H) Top-$k$ accuracy}
 \label{tab:inat_synnum_topk}
\end{table}

\subsection{Zero-Shot Recognition, CUB}
\label{sec:zsr_cub}
\remove{As previously described, there are 4 out of 200 classes in CUB which are not present in the Billow dataset.
}

\add{In addition, we compare our CE and PA proposed methods using CUB$_{\text{196}}$, which contains 196 CUB classes also contained in Billow, divided into 148 seen and 48 unseen classes. We always respect the proposed split by \cite{Yongqin2019theugly}.
Class embedding vectors were generated from illustrations using our Contrastive Encoding.
These embeddings were used in combination with TFVAEGAN~\cite{narayan2020latent},
CE-GZSL~\cite{han2021contrastive}, and LsrGAN~\cite{vyas2020leveraging} to evaluate their performance as class side information $\phi (y)$ in a ZSL setting. In Table~\ref{tab:cub_billow_results} (top) we observe that the best results with CE are obtained in combination with TFVAEGAN.
}

\add{In Table~\ref{tab:cub_billow_results} (bottom) we present an evaluation of various supervised and unsupervised domain adaptation methods for ZSL. {This was tested with DANN~\cite{dann}, MDD~\cite{mdd}, MCC~\cite{mcc}, ProtoDA~\cite{yue2021prototypical} and CCSA~\cite{motiian2017CCSA}.} 
Although DANN and ProtoDA did not completely collapse towards the seen classes, they fail to fully translate knowledge from the source domain into the target domain. 
Our PA approach on the other hand achieves the best performance, well above that of domain adaptation baselines and the CE approach.}

\add{Furthermore, we compared CE encodings of Billow illustrations with other types of side information in Table~\ref{tab:cub_others_results} using CUB$_{\text{191}}$, i.e., the subset of 191 CUB classes overlapping with other types of side-information and Billow, divided into 145 seen and 46 unseen classes.}
\remove{(1) CUB$_{\text{196}}$: the subset of 196 CUB classes also contained in Billow, divided into 148 seen and 48 unseen classes; (2) CUB$_{\text{191}}$: The subset of 191 CUB classes also present in both DNA and Billow datasets. CUB$_{\text{191}}$ contains 145 seen and 46 unseen classes.}
As in the previous experiment, the split proposed by  \cite{Yongqin2019theugly} is respected and the 
class embedding vectors were generated from illustrations using our Contrastive Encoding. We used these embeddings in combination with TFVAEGAN, CE-GZSL and LsrGAN.
We compare Billow with the following sources of side-information $\phi (y)$: binary attributes~\cite{welinder2010caltech}, visual descriptions~\cite{reed2016learning}, DNA~\cite{badirli2021fine}, and word2vec~\cite{akata2015evaluation}.
%
These experiments show that the representation power of Billow's contrastive embedding is comparable to that of word2vec and DNA embeddings.
In terms of comparison among the existing methods we can observe that TFVAEGAN achieves the best results in both scenarios. 

\begin{table}[ht]
        \begin{subtable}[t]{\linewidth}
    \caption{Experiments with Billow on CUB$_{\text{196}}$.  Top: \underline{CE Billow} (Contrastive embeddding of Billow, ours), combined with GZSL methods. Bottom: End-to-end methods to use Billow, including \underline{PA}  (Prototype Alignment, ours) and domain adaptation methods. $\dagger$ denotes UDA methods that do not use target labels}
      \resizebox{0.83\linewidth}{!}{\begin{minipage}{\linewidth}
        \include{tables/GZSL_cub_results_std_single_billow}
        \end{minipage}}
       \label{tab:cub_billow_results}
    \end{subtable}
    \begin{subtable}[t]{\linewidth}
    \caption{Experiments with Billow on CUB$_{\text{191}}$. Comparison with other types of side-information ($\phi(y)$) used with CUB.}
   \resizebox{0.83\linewidth}{!}{\begin{minipage}{\linewidth} 
\include{tables/GZSL_cub_results_std_single_others}   
        \end{minipage}}
        \label{tab:cub_others_results}
    \end{subtable}
    \caption{GZSL on CUB. Seen, unseen and harmonic mean (H) Top-$1$ accuracy. Average of 5 runs $\pm$ standard deviation. Best method for each dataset and $ \phi(y)$ is {bold}.
    }
    \label{tab:cub_all_results}
\end{table}

\subsection{Ablation Studies}

In Table~\ref{tab:ablation_losses} we show the effect of changing different components of the PA method. 
First, we observed that setting the projection head $h$ to be a small Multi Layer Perceptron decreased our performance compared to an identity function (row D). 
We speculate that the latent space of $\v{z}$ is already too close to the label domain for it to benefit from a projection head.
%
We computed $\hat{\v{y}}$ with a learned linear classifier $w_\text{cls}$ instead of using the dot product between domain embeddings and observed such modification drastically reduces the performance on the unseen classes (row A).
%
The contrastive in-domain loss $\lambda_c$ proved itself to be essential for achieving a good performance on the unseen classes (row C).
The classification loss instead seems to be more important for the recognition of the seen classes: Removing the term completely barely changes the performance on the unseen classes while drastically reducing performance on the seen ones (row B). On the other hand having a larger $\lambda_{ce}^{(t)}$ tends to boost accuracy in seen classes
with a slight reduced accuracy in unseen ones (row E).
See supplementary for an ablation with different backbones.

 \begin{table}[ht]
       \resizebox{0.77\textwidth}{!}{\begin{minipage}{\textwidth}
        \include{tables/mbank_cub_ablations_results_may} 
    \end{minipage}}
        \caption{Prototype Alignment experiments with different Hyperparameters on CUB dataset}
        \label{tab:ablation_losses}
\end{table}

%% file: tables/inat_suh_topk.tex
\begin{tabular}{ll *{3}{R{1.4cm}} l *{3}{R{1.4cm}} l *{3}{R{1.4cm}}}
\toprule
         &    & \multicolumn{3}{c}{top-1} && \multicolumn{3}{c}{top-5} && \multicolumn{3}{c}{top-10} \\\cline{3-5} \cline{7-9} \cline{11-13}
  & Model &  \multicolumn{1}{c}{S} &    \multicolumn{1}{c}{U} &    \multicolumn{1}{c}{H} && 
             \multicolumn{1}{c}{S} &    \multicolumn{1}{c}{U} &    \multicolumn{1}{c}{H} &&
             \multicolumn{1}{c}{S} &    \multicolumn{1}{c}{U} &    \multicolumn{1}{c}{H} \\ 
\multicolumn{4}{l}{iNat2017} \\\hline
 & CE &    \formatbest{33.1}~$\pm$~0.8 &     2.6~$\pm$~0.2 &    4.7~$\pm$~0.3 &&        \formatbest{57.5}~$\pm$~1.3 &        14.1~$\pm$~0.3 &        22.6~$\pm$~0.3 &&         \formatbest{66.3}~$\pm$~1.5 &         23.6~$\pm$~0.2 &         34.8~$\pm$~0.3 \\
         & PA &       23.0~$\pm$~0.3 &       \formatbest{8.8}~$\pm$~0.4 &      \formatbest{12.8}~$\pm$~0.5 &&       51.9~$\pm$~0.4 &      \formatbest{23.4}~$\pm$~0.8 &      \formatbest{32.3}~$\pm$~0.8 &&        63.8~$\pm$~0.5 &        \formatbest{32.9}~$\pm$~0.6 &        \formatbest{43.5}~$\pm$~0.6 \\
\multicolumn{4}{l}{iNat2021mini} \\\hline
 & CE &        \formatbest{24.2}~$\pm$~0.2 &         3.9~$\pm$~0.2 &    6.7~$\pm$~0.3 &&        \formatbest{46.3}~$\pm$~0.1 &        16.7~$\pm$~0.4 &        24.6~$\pm$~0.5 &&         56.4~$\pm$~0.4 &         26.5~$\pm$~0.4 &         36.1~$\pm$~0.3 \\
 & PA &       20.8~$\pm$~0.4 &      \formatbest{12.7}~$\pm$~0.4 &      \formatbest{15.7}~$\pm$~0.2 &&       46.1~$\pm$~0.5 &       \formatbest{29.0}~$\pm$~0.4 &       \formatbest{35.6}~$\pm$~0.2 &&        \formatbest{56.8}~$\pm$~0.4 &        \formatbest{38.5}~$\pm$~0.5 &  \formatbest{45.9}~$\pm$~0.3 \\
\multicolumn{4}{l}{iNat2021} \\\hline
 & CE &        \formatbest{36.6}~$\pm$~0.8 &         2.1~$\pm$~0.1 &    3.9~$\pm$~0.2 &&      \formatbest{61.1}~$\pm$~0.6 &        11.3~$\pm$~0.4 &        19.1~$\pm$~0.7 &&         \formatbest{69.7}~$\pm$~ 0.3 &         19.6~$\pm$~0.3 &         30.6~$\pm$~0.4 \\
 & PA &   20.9~$\pm$~0.3 &      \formatbest{12.2}~$\pm$~0.3 &       \formatbest{15.4}~$\pm$~0.2 &&       45.5~$\pm$~0.2 &       \formatbest{28.6}~$\pm$~0.6 &      \formatbest{35.1}~$\pm$~0.5 &&        56.6~$\pm$~0.2 &       \formatbest{37.8}~$\pm$~ 0.5 &        \formatbest{45.3}~$\pm$~0.4 \\
 
\bottomrule
\end{tabular}


%% file: tables/inat_nhops_topk.tex
\begin{tabular}{ll *{4}{R{0.91cm}} l *{4}{R{0.91cm}} l *{4}{R{0.91cm}}}
\toprule
&    & \multicolumn{4}{c}{top-1} && \multicolumn{4}{c}{top-5} && \multicolumn{4}{c}{top-10} \\ \cline{3-6} \cline{8-11} \cline{13-16}
 & Model   & 1-hop & 2-hop & 3-hop & 4-hop && 1-hop & 2-hop & 3-hop & 4-hop && 1-hop & 2-hop & 3-hop & 4-hop \\
\multicolumn{4}{l}{iNat2017} \\\hline
 & CE &    2.3 &    3.4 &    2.9 &    1.6 &&   21.3 &   16.9 &   11.4 &    7.4 &&   35.1 &   27.8 &   19.0 &   13.8 \\
         & PA &    \textbf{9.1} &    \textbf{9.9} &    \textbf{9.3} &    \textbf{7.0} &&   \textbf{29.1} &   \textbf{25.3} &   \textbf{22.2} &   \textbf{18.1} &&   \textbf{42.3} &  \textbf{ 35.4} &   \textbf{30.5} &   \textbf{25.1} \\
         \multicolumn{4}{l}{iNat2021mini} \\\hline
 & CE &    5.2 &    4.0 &    3.6 &    2.3 &&   22.9 &   16.6 &   15.1 &   10.7  &&   35.0 &   26.3 &   24.2 &   18.6 \\
         & PA &   \textbf{12.8} &   \textbf{13.6} &   \textbf{11.5} &   \textbf{11.8} &&   \textbf{33.5} &   \textbf{30.3} &   \textbf{25.9} &  \textbf{ 23.8} &&  \textbf{ 44.7} &   \textbf{40.0} &   \textbf{34.7} &   \textbf{31.2} \\
         \multicolumn{4}{l}{iNat2021} \\\hline
 & CE &    2.6 &    1.9 &    2.1 &    1.6 &&   16.5 &   11.1 &    9.4 &    7.1 &&   27.3 &   19.6 &   16.4 &   13.3 \\
         & PA &  \textbf{ 12.3} &  \textbf{ 13.3} &  \textbf{ 11.4} &  \textbf{ 10.6} &&   \textbf{33.9} &   \textbf{29.7} &  \textbf{ 25.5} &  \textbf{ 23.0} &&  \textbf{ 44.8} &  \textbf{ 39.2} &  \textbf{ 33.6} &   \textbf{30.4 }\\
\bottomrule
\end{tabular}

%% file: tables/inat_synnum_topk.tex
\begin{tabular}{ll ccc r ccc r ccc}
\toprule
         &      & \multicolumn{3}{c}{top-1} && \multicolumn{3}{c}{top-5} && \multicolumn{3}{c}{top-10} \\
         &  $N_{\text{syn}}$    &    S &   U &   H &&    S &    U &    H &&     S &    U &    H \\ \cline{3-5} \cline{7-9} \cline{11-13}
\multicolumn{6}{l}{iNat2017} \\\hline
 & 100  & \textbf{34.0} & 1.2 & 2.3 && \textbf{57.7} &  7.4 & 13.1 &&  66.0 & 13.6 & 22.5 \\
 & 1000 & 33.1 & 2.6 & 4.7 && 57.5 & 14.1 & 22.6 &&  \textbf{66.3} & 23.6 & 34.8 \\
 & 3000 & 32.8 & \textbf{3.1} & \textbf{5.7} && 56.8 & \textbf{16.1} & \textbf{25.1} &&  66.1 & \textbf{26.0} & \textbf{37.3} \\
\multicolumn{6}{l}{iNat2021mini} \\\hline
& 100  & \textbf{25.7} & 1.7 & 3.2 && \textbf{48.1} &  9.2 & 15.4 &&  \textbf{57.6} & 16.7 & 25.9 \\
& 1000 & 24.2 & 3.9 & 6.7 && 46.3 & \textbf{16.7} & 24.6 &&  56.4 & 26.5 & 36.1 \\
& 3000 & 23.1 & \textbf{4.9} & \textbf{8.0} && 45.3 & 18.6 & \textbf{26.4} &&  55.4 &\textbf{ 28.7} & \textbf{37.8} \\
\multicolumn{6}{l}{iNat2021} \\\hline        
 & 100  & \textbf{39.2} & 0.8 & 1.5 && \textbf{62.0} &  4.4 &  8.2 &&  69.5 &  7.9 & 14.2 \\
 & 1000 & 36.6 & 2.1 & 3.9 && 61.1 & 11.3 & 19.1 &&  69.7 & 19.6 & 30.6 \\
 & 3000 & 35.8 & \textbf{3.1} & \textbf{5.8} && 61.0 & \textbf{14.8} & \textbf{23.8} &&  \textbf{69.8} & \textbf{24.1} & \textbf{35.8} \\
\bottomrule
\end{tabular}

%% file: tables/GZSL_cub_results_std_single_billow.tex
\begin{tabular}{ll rrrr}
\toprule
    $\phi(y)$ & Model  & \multicolumn{1}{c}{S} & \multicolumn{1}{c}{U} & \multicolumn{1}{c}{H} \\\cline{1-5}
                 \multirow{3}{2cm}{CE Billow (ours)}  & CE-GZSL  & 42.0 $\pm$1.1 & 25.2  $\pm$1.5 & 31.5   $\pm$1.2\\
                    & LsrGAN  & \formatbest{69.7}   $\pm$0.3 &  6.4  $\pm$0.5 & 11.6  $\pm$0.9\\
                     & TFVAEGAN  & 45.5  $\pm$13.1 & \formatbest{31.5}  $\pm$5.5 & \formatbest{35.8}  $\pm$1.2\\\cline{2-5}
  \multirow{6}{2cm}{Billow (end-to-end)} &  DANN$\dagger$   &   24.3    $\pm$1.8  &   17.5  $\pm$2.3 & 20.3     $\pm$1.6 \\
                  &   MDD$\dagger$       &   1.4      $\pm$0.4  &   0.7    $\pm$0.4 & 0.9     $\pm$0.4 \\
    &           MCC$\dagger$      &   6.5     $\pm$0.5  &   5.8     $\pm$0.8  & 6.1      $\pm$0.4\\
   &            ProtoDA  & 13.8  $\pm$0.9 & 13.8  $\pm$1.8  & 14.4  $\pm$2.0\\
   &          CCSA  & \formatbest{73.5}   $\pm$0.7 &  0.1  $\pm$0.0 &  0.1  $\pm$0.1\\ 
  &        	PA (ours)  & 69.7  $\pm$0.6 & \formatbest{36.1} $\pm$1.5  &   \formatbest{47.5}$\pm$1.5 \\
\bottomrule
\end{tabular}

%% file: tables/GZSL_cub_results_std_single_others.tex
\begin{tabular}{ll rrrr}
\toprule
    $\phi(y)$ & Model  & \multicolumn{1}{c}{S} & \multicolumn{1}{c}{U} & \multicolumn{1}{c}{H} \\\cline{1-5}
  \multirow{3}{2cm}{Binary attributes} & CE-GZSL  & 59.8~$\pm$~1.9 & 48.4~$\pm$~0.7  & 53.5~$\pm$~0.7\\
                & LsrGAN   & \formatbest{63.6}~$\pm$~0.2 & 20.4~$\pm$~0.5  & 30.9~$\pm$~0.6\\
                & TFVAEGAN  & 63.4~$\pm$~2.2 & \formatbest{52.8}~$\pm$~1.4 & \formatbest{57.6}~$\pm$~0.2\\\cline{2-5}
    \multirow{3}{2cm}{Visual descriptions} & CE-GZSL  & 66.4~$\pm$~0.3 & 65.0~$\pm$~0.6 & 65.7~$\pm$~0.4 \\
                & LsrGAN  & 58.7~$\pm$~0.3 & 54.2~$\pm$~0.8 & 56.3~$\pm$~0.4 \\
                & TFVAEGAN  &\formatbest{67.8}~$\pm$~2.1 & \formatbest{68.4}~$\pm$~2.1 & \formatbest{68.1}~$\pm$~0.4 \\\cline{2-5}
    \multirow{3}{2cm}{DNA} & CE-GZSL  & 39.5 $\pm$1.2 & 13.5  $\pm$0.8 & 20.1 $\pm$0.8  \\
                & LsrGAN  & \formatbest{69.7} $\pm$0.1 & 3.9 $\pm$0.2 & 7.4  $\pm$0.4 \\
                & TFVAEGAN  & 30.8 $\pm$0.4 &\formatbest{20.3} $\pm$1.0  & \formatbest{24.5} $\pm$0.7\\\cline{2-5}
    \multirow{3}{2cm}{word2vec} & CE-GZSL  & 49.1 $\pm$1.7 & 25.9 $\pm$0.7 & 33.9 $\pm$0.5 \\
                & LsrGAN  & \formatbest{62.0}  $\pm$0.5 & 16.5  $\pm$0.4 & 26.1 $\pm$0.5\\
                & TFVAEGAN  & 45.6 $\pm$1.0 & \formatbest{27.2} $\pm$0.9  & \formatbest{34.1}  $\pm$0.9\\\cline{2-5}
    \multirow{3}{2cm}{CE Billow (ours)}  & CE-GZSL  & 42.7  $\pm$1.5 & 27.9  $\pm$0.8 & 33.8  $\pm$1.0 \\
                & LsrGAN  & \formatbest{69.2}   $\pm$0.2 &  7.0  $\pm$0.2 & 12.7  $\pm$0.4  \\
                & TFVAEGAN  & 45.3  $\pm$14.1 & \formatbest{31.6}  $\pm$5.4  & \formatbest{35.6}  $\pm$1.1 \\
\bottomrule
\end{tabular}

%% file: tables/mbank_cub_ablations_results_may.tex
\begin{tabular}{lcccccccrrr}
\toprule
& &  $\lambda_{c}$  & $h(x)$ &  $\lambda_{ce}^{t}$ &  $w_{\text{cls}}$ &  $L_{cls}(\hat{y}',y)$ &  H &    S &    U \\
\midrule
& A & 1 &     Identity &  0.1 &   learned          &               & 14.7 & 44.1 &  8.9 \\
& B &   1 &     Identity &  0.1 & $\phi(Y)^{(s)}$  &               & 42.7 & 50.6 & 37.0 \\
& C &   0 &     Identity &  0.1 &  $\phi(Y)^{(s)}$  & \checkmark    & 23.5 & 52.4 & 15.2 \\
& D &   1 &     MLP &       0.1 &  $\phi(Y)^{(s)}$   & \checkmark    & 39.3 & 50.1 & 32.3 \\
& E &   1 &     Identity &  1.0 &  $\phi(Y)^{(s)}$  & \checkmark   & 46.3 & 69.9 & 34.6 \\
& F &   {1} &     {Identity} & { 0.1} &  $\phi(Y)^{(s)}$  & \checkmark   &  47.5 & 69.7 & 36.1  \\
\bottomrule
\end{tabular}


%% file: 6_conclusions.tex
\section{Conclusion}
\label{sec:conclusions}

Our experiments show that using field guides as side information for ZSL is feasible,
expanding the set of fine-grained ZSL experiments to datasets with more natural distributions such as iNaturalist2017 and iNaturalist2021. Which are of a much larger scale than those commonly studied for ZSL (e.g., CUB, Animals with Attributes~\cite{Yongqin2019theugly}, Oxford Flowers~\cite{nilsback2008automated}).
They show that the zero-shot recognition of bird species in images is feasible with an accuracy much better than random chance. The best harmonic mean so far is obtained by the proposed PA method.
Our Experiments show that a na\"ive implementation from domain adaptation might not yield the best results, despite a comparatively small domain gap w.r.t.\ photographs. iNaturalist experiments show that, while state-of-the-art ZSL combined with the contrastive encoded illustrations achieves reasonable results, our proposed PA consistently outperforms it.
Still, identifying unseen birds across thousands of different species remains a challenge. The observed top-10 accuracies demonstrate that side-information provided by Billow indeed steers the classifier towards the correct (unseen) species. This is also reflected by the fact that we observe better performance for unseen classes that are closer to seen ones in the taxonomic hierarchy (Tab.~\ref{tab:inat_nhops_topk}).

\add{CUB has been used in combination with many types of side-information in the past. Our experiments show that leveraging illustrations in field guides can achieve comparable results to other types of side-information.} {Although attributes and keywords have higher accuracies on CUB than with Billow, illustrations are a valid alternative that contains several decades of knowledge that be realistically exploited. It appears more natural to describe new bird species using existing illustrations of them than comparing them to the test set of CUB to obtain visual descriptions~\cite{reed2016learning}, which is prone to overfit to the rather small dataset.}
%

\remove{In our work we focus on illustrations, where a na\"ive implementation from domain adaptation might not yield the best results, despite a comparatively small domain gap w.r.t.\ photographs. iNaturalist experiments show that, while state-of-the-art ZSL combined with the contrastive encoded illustrations achieves reasonable results, our proposed PA consistently outperforms it.}
%
%

Species recognition would benefit from further studies on how to incorporate more side-information, such as by explicitly modelling species similarity and patristic distances~\cite{kampffmeyer2019rethinking}; or obtain a multi-source embedding using a mixture of illustrations, text descriptions or other types of side-information. More fine-grained class representation for different sexes of the same bird species could be could further improve the results. 
While we have focused this work on illustrations of birds, there are many other field guides that could be exploited in ZSL. We hope that our work inspires more research in this direction to assist efforts in biodiversity mapping and conservation.

%% file: 7_supplementary.tex
\section{Appendix}






\subsection{Encoding of Billow}
\label{sec:exp_encoding_billow}

\begin{table}[b]
    \vspace{5pt}
\centering
\centering
    \begin{tabular}{lr}
        \toprule
        Method & \multicolumn{1}{c}{top-1} \\
\hline
        VQ-VAE  & 0.1 \\
        $\beta$-VAE & 14.2 \\
        ResNet-101$^*$ & 15.5 \\
        ResNet-50$^*$ & 12.0 \\
        ResNet-18$^*$ & 16.5\\
        ResNet-18 (ours) & \textbf{17.7} \\
        \bottomrule
    \end{tabular}
        \caption{Top-1 species accuracy on a test set of Billow samples with different encoders. 
    $^*$ indicates models without fine-tuning 
    }
\label{tab:billow_encoding_results}
\end{table}

Appropriate encoding of the side information can have a strong effect in the overall performance.
We evaluated different ways to encode our illustrations to be used with ZSL state-of-the-art methods.
From Billow, we created a separate test set and evaluated the predictive power for species classification of each encoding. This is a challenging task, since we only have a couple of illustrations (one male; one female; and , in some cases, a head detail) per species. In contrast, traditional methos usually require a large amount of images per species to train be able to perform automatic classification.

To evaluate the quality of our encoding method we measured how well can we predict the class at different hierarchical levels. Out of the 10'631 we take all the species whose genus has at least 5 species in the dataset and create a train/validation split from them. This results in 18'489 illustrations: 13362 for training, 1908 for validation and 3219 for testing. In all splits  \add{combined} there are samples of 8646 species, 956 genus, 175 families and 33 orders. We evaluate top-1 and top-10 accuracy on all 4 hierarchical.

For training we explored Variational Auto-Encoder (VAE) generative models to encode our dataset. VAEs consist of two networks, an encoder $E$ and a decoder $D$. A regular auto-encoder would simply use the output of the encoder and feed it to the decoder to reconstruct the input. For training the auto-encoder, the reconstruction loss is defined as $L = d(x, D(E(x)))$, where  $d$ is usually a Euclidean 
 distance between the input and its reconstruction. VAEs assume a prior on the output of the encoder $p_z$ and maximize the log-likelihood of the reconstruction produced by $D$ over the entire prior distribution $p_z$.

Modelling such a distribution would be desirable in our case as we will use the embedding and the distances between them for ZSL. Hence we explore 
two VAE variants: $\beta$-VAE~\cite{Higgins2017betaVAELB,burgess2018understanding} and VQ-VAE~\cite{van2017neural,razavi2019generating}. Our motivation to test these 2 variations of VAE is to explore the effect of different priors on the latent distribution $p \sim z$. $\beta$-VAE uses a more constrained information bottleneck on the embedding $z$ than vanilla-VAE (i.e. $\beta > 1$) to obtain a disentangled representation $z$. As a reconstruction, we slowly increase the bottleneck capacity over training as proposed by \cite{burgess2018understanding}. VQ-VAE on other hand imposes a discrete distribution over the embedding $z$, this allows to control the information bottle neck by imposing a very small dimension on the discrete distribution.


Additionally we fine-tuned a ResNet classifier pretrained on ImageNet. The classifier was supervised by $L = L_{cls} + L_{cont}$, where $L_{cont}$ is as defined in Eq.~\ref{eq:contrastive_loss_encoded}.
VAE experiments were trained their corresponding reconstruction loss and the supervision loss $L_{cls}$.

Once the model was trained we evaluated its predictive power by feeding the embedding $z$ into a small Multi-layer Perceptron network and trained to predict the level of the label $k$. The results on the test set in Table~\ref{tab:billow_encoding_results} show that $\beta$-VAE does not perform close to the ResNet models. Along with our fine-tuned ResNet-18, we include results on pretrained modesl on Image-Net without fine-tuning the backbone. As expected larger networks achieved higher performance, but our fine-tuning on ResNet-18 improved performance in the most challenging case of fine-grained species recognition. 
Our ResNet-18 already achieved 100\% accuracy on the training set and observed over-fitting on the validation set after fine-tuning. For this reason we decided to keep the fine-tuned ResNet18 as the default encoder for billow.

\begin{table}[t]
      \resizebox{0.87\linewidth}{!}{\begin{minipage}{\linewidth}
      \centering
        \include{tables/inat_level_topk}  
        \end{minipage}}
\caption{Unseen $n$-hop validation sets top-$1$ accuracy at different label hierarchy levels. Average of 5 runs $\pm$ standard deviation. }
        \label{tab:inat_level_topk}
\end{table}

\begin{table*}[t]
    \caption{\add{Prototype Alignment and Domain adaptation baseline experiments with different ResNet backbones on iNaturalist. Top-$k$ Accuracies}}
      \resizebox{0.9\textwidth}{!}{\begin{minipage}{\textwidth}
       \centering
        \include{tables/mbank_inat_backbones_results-may_std} 
        \end{minipage}}
        \label{tab:ablation_backbones_inat}
\end{table*}

\begin{table}[t]
      \centering
      \resizebox{0.86\linewidth}{!}{\begin{minipage}{\linewidth}
      \centering
        \include{tables/mbank_cub_backbones_results-may_std}  
        \end{minipage}}
    \caption{Prototype Alignment and Domain adaptation baselines experiments with different ResNet backbones on CUB Dataset. Top-1 Accuracy}
        \label{tab:ablation_backbones_cub}
\end{table}

\subsection{Accuracy at different hierarchical levels}
\label{sec:accuracy_at_level}
We evaluated top-1 accuracy at different label hierarchies on each n-hop validation set (which were created using the label hierarchies, see Section~\ref{sec:ilustrations_zsl} for details). A prediction is correct at the family-level if the predicted species was of the same family as the target species. We present these results in Table~\ref{tab:inat_level_topk}. The 2-hop (set where no species of the same family are part of the seen species) performance between species and genus is similar, suggesting that the Zero-shot task is equally difficult at the 2 considered hierarchy levels. Similarly for 4-hop (no overlap at any hierarchy label), the performance is equally low at any level. These results suggest that the label hierarchical distance is a meaningful strategy for evaluation. Future work exploit this label hierarchy at training time.

\subsection{Ablation: Backbone Size}

\add{We explore the effect of using different backbones using end-to-end methods, including different Domain Adaptation Baselines and PA (ours).
For iNaturalist datasets in Table~\ref{tab:ablation_backbones_inat}, we observe better accuracies with larger networks, and slightly higher accuracies with ResNet-101. For iNat2021mini, aligned with what we observed before, we observe better performance than iNat2020 in all cases considered in these experiments.
The results on CUB dataset in Table~\ref{tab:ablation_backbones_cub} show that the performance increases for all the baselines using larger ResNets, while some of the methods have slightly higher accuracies with ResNet-50 the difference are within the marging of error compared to ResNet-101.
}

%% file: tables/inat_level_topk.tex
\begin{tabular}{lllrrrr}
\toprule
 & Level & \multicolumn{1}{c}{1-hop}  &   \multicolumn{1}{c}{2-hop}  &  \multicolumn{1}{c}{3-hop}    &  \multicolumn{1}{c}{4-hop}  
\\ \cline{3-6}
\multicolumn{4}{l}{iNat2017} \\\hline
  & species &  9.1 $\pm$ 0.4 &  9.9 $\pm$ 0.5 &  9.3 $\pm$ 0.4 &  7.0 $\pm$ 0.7 \\
           & genus & 29.4 $\pm$ 0.5 & 10.2 $\pm$ 0.5 & 12.9 $\pm$ 0.6 & 11.0 $\pm$ 1.4 \\
            & family & 50.2 $\pm$ 0.7 & 36.7 $\pm$ 0.9 & 12.9 $\pm$ 0.6 & 16.8 $\pm$ 2.6 \\
            & order & 77.8 $\pm$ 0.7 & 67.7 $\pm$ 0.6 & 69.7 $\pm$ 1.4 & 16.8 $\pm$ 2.6 \\
\multicolumn{4}{l}{iNat2021mini} \\\hline
 & species & 12.8 $\pm$ 0.5 & 13.6 $\pm$ 0.4 & 11.4 $\pm$ 0.8 & 11.6 $\pm$ 0.4 \\
            & genus & 29.3 $\pm$ 0.4 & 13.7 $\pm$ 0.4 & 17.1 $\pm$ 1.1 & 16.2 $\pm$ 0.7 \\
             & family & 47.4 $\pm$ 0.8 & 38.3 $\pm$ 0.4 & 17.4 $\pm$ 1.2 & 20.6 $\pm$ 1.1 \\
             & order & 75.1 $\pm$ 1.0 & 69.9 $\pm$ 1.0 & 69.0 $\pm$ 1.3 & 20.6 $\pm$ 1.1 \\
\multicolumn{4}{l}{iNat2021} \\\hline
  & species & 12.3 $\pm$ 0.3 & 13.4 $\pm$ 0.6 & 11.2 $\pm$ 0.3 & 10.6 $\pm$ 0.4 \\
             & genus & 28.8 $\pm$ 0.7 & 13.5 $\pm$ 0.6 & 16.8 $\pm$ 0.6 & 15.7 $\pm$ 0.7 \\
             & family & 46.8 $\pm$ 0.5 & 37.8 $\pm$ 0.6 & 16.9 $\pm$ 0.6 & 20.7 $\pm$ 0.7 \\
             & order & 74.7 $\pm$ 0.4 & 70.1 $\pm$ 0.8 & 69.8 $\pm$ 0.8 & 20.7 $\pm$ 0.7 \\
\bottomrule
\end{tabular}

%% file: tables/mbank_inat_backbones_results-may_std.tex

\begin{tabular}{ll *{3}{R{1.4cm}} l *{3}{R{1.4cm}} l *{3}{R{1.4cm}}}
\toprule
         &    & \multicolumn{3}{c}{top-1} && \multicolumn{3}{c}{top-5} && \multicolumn{3}{c}{top-10} \\\cline{3-5} \cline{7-9} \cline{11-13}
  & Backbone &  \multicolumn{1}{c}{S} &    \multicolumn{1}{c}{U} &    \multicolumn{1}{c}{H} && 
             \multicolumn{1}{c}{S} &    \multicolumn{1}{c}{U} &    \multicolumn{1}{c}{H} &&
             \multicolumn{1}{c}{S} &    \multicolumn{1}{c}{U} &    \multicolumn{1}{c}{H} \\ 
\multicolumn{4}{l}{iNat2017} \\\hline
           & ResNet-18 &       13.1$\pm$0.3 &        7.6$\pm$0.5 &        9.6$\pm$0.3 &&       34.4$\pm$0.6 &       20.6$\pm$0.6 &       25.8$\pm$0.4 &&        46.0$\pm$0.7 &        28.8$\pm$0.5 &        35.4$\pm$0.4 \\
           & ResNet-50 &       20.9$\pm$0.3 &        8.2$\pm$0.3 &       11.8$\pm$0.4 &&       48.8$\pm$0.6 &       21.8$\pm$0.7 &       30.1$\pm$0.7 &&        60.9$\pm$0.6 &        30.9$\pm$0.5 &        41.0$\pm$0.5 \\
 & ResNet-101 &       \textbf{23.0}$\pm$0.3 &        \textbf{8.8}$\pm$0.4 &       \textbf{12.8}$\pm$0.5 &&       \textbf{51.9}$\pm$0.4 &       \textbf{23.4}$\pm$0.8 &       \textbf{32.3}$\pm$0.8 &&        \textbf{63.8}$\pm$0.5 &        \textbf{32.9}$\pm$0.6 &        \textbf{43.5}$\pm$0.6 \\
\multicolumn{4}{l}{iNat2021} \\\hline
           & ResNet-18 &       11.1$\pm$0.2 &        7.9$\pm$0.2 &        9.3$\pm$0.1 &&       29.2$\pm$0.2 &       19.8$\pm$0.3 &       23.6$\pm$0.2 &&        39.7$\pm$0.4 &        27.4$\pm$0.4 &        32.4$\pm$0.3 \\
           & ResNet-50 &       17.9$\pm$0.5 &       10.5$\pm$0.2 &       13.2$\pm$0.3 &&       41.4$\pm$0.5 &       24.8$\pm$0.5 &       31.0$\pm$0.5 &&        53.2$\pm$0.5 &        33.8$\pm$0.5 &        41.3$\pm$0.5 \\
 & ResNet-101 &       \textbf{20.9}$\pm$0.3 &       \textbf{12.2}$\pm$0.3 &       \textbf{15.4}$\pm$0.2 &&       \textbf{45.5}$\pm$0.2 &       \textbf{28.6}$\pm$0.6 &       \textbf{35.1}$\pm$0.5 &&        \textbf{56.6}$\pm$0.2 &        \textbf{37.8}$\pm$0.5 &        \textbf{45.3}$\pm$0.4 \\
\multicolumn{4}{l}{iNat2021mini} \\\hline
           & ResNet-18 &       11.2$\pm$0.3 &        8.3$\pm$0.2 &        9.5$\pm$0.1 &&       29.4$\pm$0.3 &       20.6$\pm$0.3 &       24.2$\pm$0.2 &&        40.0$\pm$0.3 &        28.5$\pm$0.3 &        33.3$\pm$0.1 \\
           & ResNet-50 &       18.3$\pm$0.4 &       10.8$\pm$0.3 &       13.6$\pm$0.3 &&       41.9$\pm$0.5 &       25.5$\pm$0.5 &       31.7$\pm$0.5 &&        53.5$\pm$0.3 &        34.3$\pm$0.7 &        41.8$\pm$0.5 \\
 & ResNet-101 &       \textbf{20.8}$\pm$0.4 &       \textbf{12.7}$\pm$0.4 &       \textbf{15.7}$\pm$0.2 &&       \textbf{46.1}$\pm$0.5 &       \textbf{29.0}$\pm$0.4 &       \textbf{35.6}$\pm$0.2 &&        \textbf{56.8}$\pm$0.4 &        \textbf{38.5}$\pm$0.5 &        \textbf{45.9}$\pm$0.3 \\
\bottomrule
\end{tabular}

%% file: tables/mbank_cub_backbones_results-may_std.tex
\begin{tabular}{llrrr}
    	\toprule
       & Model  & \multicolumn{1}{c}{S} & \multicolumn{1}{c}{U} & \multicolumn{1}{c}{H} \\\cline{3-5}
         \multicolumn{4}{l}{Resnet-18} \\    \hline
         & DANN~\cite{dann} & 13.0 $\pm$ 1.1 & 10.3 $\pm$ 0.5  & 11.5 $\pm$ 0.5 \\
         {}  & MDD~\cite{mdd}   &  1.2 $\pm$ 0.2 & 0.0 $\pm$ 0.1  & 0.0 $\pm$ 0.1 \\
         {}    & MCC~\cite{mcc}  & 4.4 $\pm$ 0.5 & 3.8 $\pm$ 0.4   & 4.0 $\pm$ 0.2 \\
        & PA  & 48.2 $\pm$0.4 & 37.2  $\pm$0.4 & 42.0  $\pm$0.2 \\
               \multicolumn{4}{l}{Resnet-50} \\   \hline
               & DANN~\cite{dann} & 16.3 $\pm$ 0.6  & 14.4 $\pm$ 1.5  & 15.2 $\pm$ 1.1 \\
        {}  & MDD~\cite{mdd}    & 0.6 $\pm$ 0.3 & 0.9 $\pm$ 1.2  & 0.2 $\pm$ 0.5 \\
        {}          & MCC~\cite{mcc}   & 5.8 $\pm$ 0.2 & 6.7 $\pm$ 0.7  & 6.2 $\pm$ 0.4 \\
& PA  & 64.8 $\pm$0.5 & \textbf{37.8} $\pm$1.0 & \textbf{47.8} $\pm$0.8 \\
               \multicolumn{4}{l}{Resnet-101} \\   \hline
    & DANN~\cite{dann}  & 24.3 $\pm$ 1.8  & 17.6 $\pm$ 2.4  & 20.3 $\pm$ 1.6 \\
        {} & MDD~\cite{mdd}   & 1.4 $\pm$ 0.4 & 0.7 $\pm$ 0.5 & 0.9 $\pm$ 0.4  \\
        {}          & MCC~\cite{mcc}   & 6.6 $\pm$ 0.5  & 5.8 $\pm$ 0.6   & 6.1 $\pm$ 0.4 \\
        & PA & \textbf{69.9}   $\pm$ 0.6 & 34.6  $\pm$1.5 & 46.3   $\pm$1.5 \\
     \bottomrule
\end{tabular}

%% file: main_wacv.bbl
\begin{thebibliography}{10}\itemsep=-1pt

\bibitem{akata2013label}
Zeynep Akata, Florent Perronnin, Zaid Harchaoui, and Cordelia Schmid.
\newblock Label-embedding for attribute-based classification.
\newblock In {\em IEEE Conference on Computer Vision and Pattern Recognition},
  2013.

\bibitem{akata2015label}
Zeynep Akata, Florent Perronnin, Zaid Harchaoui, and Cordelia Schmid.
\newblock Label-embedding for image classification.
\newblock {\em IEEE Transactions on Pattern Analysis and Machine Intelligence},
  38(7):1425--1438, 2015.

\bibitem{akata2015evaluation}
Zeynep Akata, Scott Reed, Daniel Walter, Honglak Lee, and Bernt Schiele.
\newblock Evaluation of output embeddings for fine-grained image
  classification.
\newblock In {\em IEEE Conference on Computer Vision and Pattern Recognition},
  2015.

\bibitem{antol2014zero}
Stanislaw Antol, C~Lawrence Zitnick, and Devi Parikh.
\newblock Zero-shot learning via visual abstraction.
\newblock In {\em European Conference on Computer Vision}, 2014.

\bibitem{badirli2021fine}
Sarkhan Badirli, Zeynep Akata, George Mohler, Christine Picard, and Mehmet
  Dundar.
\newblock Fine-grained zero-shot learning with dna as side information.
\newblock {\em Advances in Neural Information Processing Systems}, 2021.

\bibitem{barz2020deep}
Bjorn Barz and Joachim Denzler.
\newblock Deep learning on small datasets without pre-training using cosine
  loss.
\newblock In {\em IEEE/CVF Winter Conference on Applications of Computer
  Vision}, 2020.

\bibitem{billerman2020birds}
SM Billerman, BK Keeney, PG Rodewald, and TS Schulenberg.
\newblock Birds of the world. cornell laboratory of ornithology, 2020.

\bibitem{burgess2018understanding}
Christopher~P Burgess, Irina Higgins, Arka Pal, Loic Matthey, Nick Watters,
  Guillaume Desjardins, and Alexander Lerchner.
\newblock Understanding disentangling in $\beta$-{VAE}.
\newblock {\em arXiv preprint arXiv:1804.03599}, 2018.

\bibitem{daume2007frustratingly}
H Daume.
\newblock Frustratingly easy domain adaptation.
\newblock In {\em Annual Meeting of the Association for Computational
  Linguistics, 2007}, 2007.

\bibitem{ipbes}
S D{\'i}az, J Settele, E Brond{\'i}zio, H Ngo, M Gu{\`e}ze, J Agard, A Arneth,
  P Balvanera, K Brauman, S Butchart, K Chan, L Garibaldi, K Ichii, J Liu, S
  Subramanian, G Midgley, P Miloslavich, Z Moln{\'a}r, D Obura, A Pfaff, S
  Polasky, A Purvis, Jona Razzaque, B Reyers, R Chowdhury, Y Shin, I
  {Visseren-Hamakers}, K Willis, and C Zayas.
\newblock Summary for policymakers of the global assessment report on
  biodiversity and ecosystem services.
\newblock Technical report, Intergovernmental Science-Policy Platform on
  Biodiversity and Ecosystem Services, 2019.

\bibitem{fagerlund2007bird}
Seppo Fagerlund.
\newblock Bird species recognition using support vector machines.
\newblock {\em EURASIP Journal on Advances in Signal Processing}, 2007:1--8,
  2007.

\bibitem{farnsworth2013next}
Elizabeth~J. Farnsworth, Miyoko Chu, W.~John Kress, Amanda~K. Neill, Jason~H.
  Best, John Pickering, Robert~D. Stevenson, Gregory~W. Courtney, John~K.
  VanDyk, and Aaron~M. Ellison.
\newblock {Next-Generation Field Guides}.
\newblock {\em BioScience}, 63(11):891--899, 2013.

\bibitem{frome2013devise}
Andrea Frome, Greg~S Corrado, Jon Shlens, Samy Bengio, Jeff Dean, Marc'Aurelio
  Ranzato, and Tomas Mikolov.
\newblock Devise: A deep visual-semantic embedding model.
\newblock {\em Advances in Neural Information Processing Systems}, 2013.

\bibitem{FromeDeViseNIPS2013}
Andrea Frome, Greg~S Corrado, Jon Shlens, Samy Bengio, Jeff Dean,
  Marc\textquotesingle~Aurelio Ranzato, and Tomas Mikolov.
\newblock {DeViSE}: A deep visual-semantic embedding model.
\newblock In {\em Advances in Neural Information Processing Systems}, 2013.

\bibitem{grl}
Yaroslav Ganin and Victor Lempitsky.
\newblock Unsupervised domain adaptation by backpropagation.
\newblock In {\em International Conference on Machine Learning}, 2015.

\bibitem{dann}
Yaroslav Ganin, Evgeniya Ustinova, Hana Ajakan, Pascal Germain, Hugo
  Larochelle, Fran{\c{c}}ois Laviolette, Mario March, and Victor Lempitsky.
\newblock Domain-adversarial training of neural networks.
\newblock {\em Journal of Machine Learning Research}, 17(59):1--35, 2016.

\bibitem{han2021contrastive}
Zongyan Han, Zhenyong Fu, Shuo Chen, and Jian Yang.
\newblock Contrastive embedding for generalized zero-shot learning.
\newblock In {\em IEEE/CVF Conference on Computer Vision and Pattern
  Recognition}, 2021.

\bibitem{He_2016_CVPR}
Kaiming He, Xiangyu Zhang, Shaoqing Ren, and Jian Sun.
\newblock Deep residual learning for image recognition.
\newblock In {\em IEEE Conference on Computer Vision and Pattern Recognition},
  2016.

\bibitem{Higgins2017betaVAELB}
Irina Higgins, Lo{\"i}c Matthey, Arka Pal, Christopher~P. Burgess, Xavier
  Glorot, Matthew~M. Botvinick, Shakir Mohamed, and Alexander Lerchner.
\newblock beta-{VAE}: Learning basic visual concepts with a constrained
  variational framework.
\newblock In {\em International Conference on Learning Representations}, 2017.

\bibitem{iNat_web}
{iNaturalist}.
\newblock \url{https://www.inaturalist.org}.
\newblock California Academy of Sciences \& National Geographic Society, 2011.
\newblock Accessed: 26-05-2021.

\bibitem{mcc}
Ying Jin, Ximei Wang, Mingsheng Long, and Jianmin Wang.
\newblock Minimum class confusion for versatile domain adaptation.
\newblock In {\em European Conference on Computer Vision}, 2020.

\bibitem{kampffmeyer2019rethinking}
Michael Kampffmeyer, Yinbo Chen, Xiaodan Liang, Hao Wang, Yujia Zhang, and
  Eric~P Xing.
\newblock Rethinking knowledge graph propagation for zero-shot learning.
\newblock In {\em IEEE/CVF Conference on Computer Vision and Pattern
  Recognition}, 2019.

\bibitem{Prannay2020supervcontr}
Prannay Khosla, Piotr Teterwak, Chen Wang, Aaron Sarna, Yonglong Tian, Phillip
  Isola, Aaron Maschinot, Ce Liu, and Dilip Krishnan.
\newblock Supervised contrastive learning.
\newblock In {\em Advances in Neural Information Processing Systems}, 2020.

\bibitem{kim2020cross}
Donghyun Kim, Kuniaki Saito, Tae-Hyun Oh, Bryan~A Plummer, Stan Sclaroff, and
  Kate Saenko.
\newblock Cross-domain self-supervised learning for domain adaptation with few
  source labels.
\newblock {\em arXiv preprint arXiv:2003.08264}, 2020.

\bibitem{kingma2014adam}
Diederik~P Kingma and Jimmy Ba.
\newblock Adam: A method for stochastic optimization.
\newblock {\em arXiv preprint arXiv:1412.6980}, 2014.

\bibitem{Lambert2009learning}
Christoph~H. Lampert, Hannes Nickisch, and Stefan Harmeling.
\newblock Learning to detect unseen object classes by between-class attribute
  transfer.
\newblock In {\em IEEE Conference on Computer Vision and Pattern Recognition},
  2009.

\bibitem{larochelle2008zero}
Hugo Larochelle, Dumitru Erhan, and Yoshua Bengio.
\newblock Zero-data learning of new tasks.
\newblock In {\em AAAI Conference on Artificial Intelligence}, 2008.

\bibitem{motiian2017CCSA}
Saeid Motiian, Marco Piccirilli, Donald~A. Adjeroh, and Gianfranco Doretto.
\newblock Unified deep supervised domain adaptation and generalization.
\newblock In {\em IEEE International Conference on Computer Vision}, 2017.

\bibitem{narayan2020latent}
Sanath Narayan, Akshita Gupta, Fahad~Shahbaz Khan, Cees~GM Snoek, and Ling
  Shao.
\newblock Latent embedding feedback and discriminative features for zero-shot
  classification.
\newblock In {\em European Conference on Computer Vision}, 2020.

\bibitem{nilsback2008automated}
Maria-Elena Nilsback and Andrew Zisserman.
\newblock Automated flower classification over a large number of classes.
\newblock In {\em 2008 Sixth Indian Conference on Computer Vision, Graphics \&
  Image Processing}, pages 722--729. IEEE, 2008.

\bibitem{norouzi2013zero}
Mohammad Norouzi, Tomas Mikolov, Samy Bengio, Yoram Singer, Jonathon Shlens,
  Andrea Frome, Greg~S Corrado, and Jeffrey Dean.
\newblock Zero-shot learning by convex combination of semantic embeddings.
\newblock {\em arXiv preprint arXiv:1312.5650}, 2013.

\bibitem{paszke2017automatic}
Adam Paszke, Sam Gross, Soumith Chintala, Gregory Chanan, Edward Yang, Zachary
  DeVito, Zeming Lin, Alban Desmaison, Luca Antiga, and Adam Lerer.
\newblock Automatic differentiation in {PyTorch}.
\newblock In {\em Advances in Neural Information Processing Systems,
  Workshops}, 2017.

\bibitem{priyadarshani2020wavelet}
Nirosha Priyadarshani, Stephen Marsland, Julius Juodakis, Isabel Castro, and
  Virginia Listanti.
\newblock Wavelet filters for automated recognition of birdsong in long-time
  field recordings.
\newblock {\em Methods in Ecology and Evolution}, 11(3):403--417, 2020.

\bibitem{razavi2019generating}
Ali Razavi, Aaron Van~den Oord, and Oriol Vinyals.
\newblock Generating diverse high-fidelity images with vq-vae-2.
\newblock {\em Advances in Neural Information Processing Systems}, 2019.

\bibitem{reed2016learning}
Scott Reed, Zeynep Akata, Honglak Lee, and Bernt Schiele.
\newblock Learning deep representations of fine-grained visual descriptions.
\newblock In {\em IEEE Conference on Computer Vision and Pattern Recognition},
  2016.

\bibitem{sangkloy2016sketchy}
Patsorn Sangkloy, Nathan Burnell, Cusuh Ham, and James Hays.
\newblock The sketchy database: learning to retrieve badly drawn bunnies.
\newblock {\em ACM Transactions on Graphics}, 35(4):1--12, 2016.

\bibitem{shen2020invertible}
Yuming Shen, Jie Qin, Lei Huang, Li Liu, Fan Zhu, and Ling Shao.
\newblock Invertible zero-shot recognition flows.
\newblock In {\em European Conference on Computer Vision}, 2020.

\bibitem{shigeto2015ridge}
Yutaro Shigeto, Ikumi Suzuki, Kazuo Hara, Masashi Shimbo, and Yuji Matsumoto.
\newblock Ridge regression, hubness, and zero-shot learning.
\newblock In {\em Machine Learning and Knowledge Discovery in Databases}, 2015.

\bibitem{stowell2019automatic}
Dan Stowell, Michael~D Wood, Hanna Pamu{\l}a, Yannis Stylianou, and Herv{\'e}
  Glotin.
\newblock Automatic acoustic detection of birds through deep learning: the
  first bird audio detection challenge.
\newblock {\em Methods in Ecology and Evolution}, 10(3):368--380, 2019.

\bibitem{sullivan2009ebird}
Brian~L Sullivan, Christopher~L Wood, Marshall~J Iliff, Rick~E Bonney, Daniel
  Fink, and Steve Kelling.
\newblock {eBird}: A citizen-based bird observation network in the biological
  sciences.
\newblock {\em Biological Conservation}, 142(10):2282--2292, 2009.

\bibitem{coral}
Baochen Sun, Jiashi Feng, and Kate Saenko.
\newblock Return of frustratingly easy domain adaptation.
\newblock In {\em AAAI Conference on Artificial Intelligence}, 2016.

\bibitem{deep_coral}
Baochen Sun and Kate Saenko.
\newblock Deep {CORAL}: Correlation alignment for deep domain adaptation.
\newblock In {\em European Conference on Computer Vision Workshops}, 2016.

\bibitem{adda}
Eric Tzeng, Judy Hoffman, Kate Saenko, and Trevor Darrell.
\newblock Adversarial discriminative domain adaptation.
\newblock In {\em IEEE Conference on Computer Vision and Pattern Recognition},
  2017.

\bibitem{van2017neural}
Aaron Van Den~Oord, Oriol Vinyals, et~al.
\newblock Neural discrete representation learning.
\newblock {\em Advances in Neural Information Processing Systems}, 2017.

\bibitem{van2021benchmarking}
Grant {Van Horn}, Elijah Cole, Sara Beery, Kimberly Wilber, Serge Belongie, and
  Oisin {Mac Aodha}.
\newblock Benchmarking representation learning for natural world image
  collections.
\newblock In {\em IEEE/CVF Conference on Computer Vision and Pattern
  Recognition}, 2021.

\bibitem{van2018inaturalist}
Grant Van~Horn, Oisin Mac~Aodha, Yang Song, Yin Cui, Chen Sun, Alex Shepard,
  Hartwig Adam, Pietro Perona, and Serge Belongie.
\newblock The {iNaturalist} species classification and detection dataset.
\newblock In {\em IEEE Conference on Computer Vision and Pattern Recognition},
  2018.

\bibitem{vyas2020leveraging}
Maunil~R. Vyas, Hemanth Venkateswara, and Sethuraman Panchanathan.
\newblock Leveraging seen and unseen semantic relationships for generative
  zero-shot learning.
\newblock In {\em European Conference on Computer Vision}, 2020.

\bibitem{few_shot_survey}
Yaqing Wang, Quanming Yao, James~T. Kwok, and Lionel~M. Ni.
\newblock Generalizing from a few examples: A survey on few-shot learning.
\newblock {\em ACM Computing Surveys}, 53(3), 2020.

\bibitem{welinder2010caltech}
P. Welinder, S. Branson, T. Mita, C. Wah, F. Schroff, S. Belongie, and P.
  Perona.
\newblock {Caltech-UCSD Birds 200}.
\newblock Technical Report CNS-TR-2010-001, California Institute of Technology,
  2010.

\bibitem{Yongqin2019theugly}
Yongqin Xian, Christoph~H. Lampert, Bernt Schiele, and Zeynep Akata.
\newblock Zero-shot learning -- a comprehensive evaluation of the good, the bad
  and the ugly.
\newblock {\em IEEE Transactions on Pattern Analysis and Machine Intelligence},
  41(9):2251--2265, 2019.

\bibitem{xian2018feature}
Yongqin Xian, Tobias Lorenz, Bernt Schiele, and Zeynep Akata.
\newblock Feature generating networks for zero-shot learning.
\newblock In {\em IEEE Conference on Computer Vision and Pattern Recognition},
  2018.

\bibitem{yue2021prototypical}
Xiangyu Yue, Zangwei Zheng, Shanghang Zhang, Yang Gao, Trevor Darrell, Kurt
  Keutzer, and Alberto~Sangiovanni Vincentelli.
\newblock Prototypical cross-domain self-supervised learning for few-shot
  unsupervised domain adaptation.
\newblock In {\em IEEE/CVF Conference on Computer Vision and Pattern
  Recognition}, 2021.

\bibitem{yue2021counterfactual}
Zhongqi Yue, Tan Wang, Qianru Sun, Xian-Sheng Hua, and Hanwang Zhang.
\newblock Counterfactual zero-shot and open-set visual recognition.
\newblock In {\em IEEE/CVF Conference on Computer Vision and Pattern
  Recognition}, 2021.

\bibitem{zhang2017learning}
Li Zhang, Tao Xiang, and Shaogang Gong.
\newblock Learning a deep embedding model for zero-shot learning.
\newblock In {\em IEEE Conference on Computer Vision and Pattern Recognition},
  2017.

\bibitem{mdd}
Yuchen Zhang, Tianle Liu, Mingsheng Long, and Michael Jordan.
\newblock Bridging theory and algorithm for domain adaptation.
\newblock In {\em International Conference on Machine Learning}, 2019.

\end{thebibliography}
